\documentclass[10pt,twocolumn,letterpaper]{article}

\usepackage[pagenumbers]{cvpr} 

\definecolor{cvprblue}{rgb}{0.21,0.49,0.74}
\usepackage[pagebackref,breaklinks,colorlinks,allcolors=cvprblue]{hyperref}

\usepackage[utf8]{inputenc} 
\usepackage[T1]{fontenc}    
\usepackage{url}            
\usepackage{booktabs}       
\usepackage{amsfonts}       
\usepackage{nicefrac}       
\usepackage{microtype}      
\usepackage{xcolor}         
\usepackage[utf8]{inputenc} 
\usepackage[T1]{fontenc}    
\usepackage{hyperref}
\hypersetup{colorlinks}
\usepackage{url}           
\usepackage{booktabs}       
\usepackage{amsmath}       
\usepackage{nicefrac}     
\usepackage{microtype}      
\usepackage{multirow}      
\usepackage{graphicx}
\usepackage{graphics}
\usepackage{tabularx}
\usepackage{makecell}
\usepackage{caption}
\usepackage{wrapfig}
\usepackage{enumitem}
\usepackage{subcaption}
\usepackage{amsfonts}      
\usepackage{nicefrac}
\usepackage{amsmath,bm}
\usepackage{enumitem}
\usepackage{algorithmic}
\usepackage{algorithm}
\usepackage{verbatim}
\usepackage{listings}
\usepackage{wrapfig}
\usepackage{mathtools}
\definecolor{codeblue}{rgb}{0.25,0.5,0.25}
\definecolor{codekw}{rgb}{0.85, 0.18, 0.50}
\usepackage{subcaption}
\usepackage{tikz}
\usepackage{comment}
\usepackage{overpic}
\newcommand{\smallnamefont}{\fontsize{4.6pt}{4.6pt}\selectfont} 

\def\vs{{\em vs.~}}

\usepackage{color, colortbl}
\definecolor{Gray}{gray}{0.93}
\newcommand{\tablestyle}[2]{\setlength{\tabcolsep}{#1}\renewcommand{\arraystretch}{#2}\centering\scriptsize}

\definecolor{Gray}{gray}{0.93}

\usepackage[margin=4pt,font=small,labelfont=bf,labelsep=endash,tableposition=bottom]{caption}
\newcolumntype{H}{>{\setbox0=\hbox\bgroup}c<{\egroup}@{}}

\def\paperID{65} 
\def\confName{CVPR}
\def\confYear{2025}

\title{MambaVision: A Hybrid Mamba-Transformer Vision
Backbone}

\author{Ali Hatamizadeh, Jan Kautz \vspace{1mm} \\ 
NVIDIA\\
{\tt\small \{ahatamizadeh, jkautz\}@nvidia.com}
}
\begin{document}
\maketitle

\begin{abstract}
We propose a novel hybrid Mamba-Transformer backbone, MambaVision, specifically tailored for vision applications. Our core contribution includes redesigning the Mamba formulation to enhance its capability for efficient modeling of visual features. Through a comprehensive ablation study, we demonstrate the feasibility of integrating Vision Transformers (ViT) with Mamba. Our results show that equipping the Mamba architecture with self-attention blocks in the final layers greatly improves its capacity to capture long-range spatial dependencies. Based on these findings, we introduce a family of MambaVision models with a hierarchical architecture to meet various design criteria. For classification on the ImageNet-1K dataset, MambaVision variants achieve state-of-the-art (SOTA) performance in terms of both Top-1 accuracy and throughput. In downstream tasks such as object detection, instance segmentation, and semantic segmentation on MS COCO and ADE20K datasets, MambaVision outperforms comparably sized backbones while demonstrating favorable performance. Code:  \url{https://github.com/NVlabs/MambaVision}  
\end{abstract}

\begin{figure}[t] 
\small
\centering
    \begin{minipage}[c]{\linewidth}
    \tiny
  \begin{overpic}[width=\textwidth]{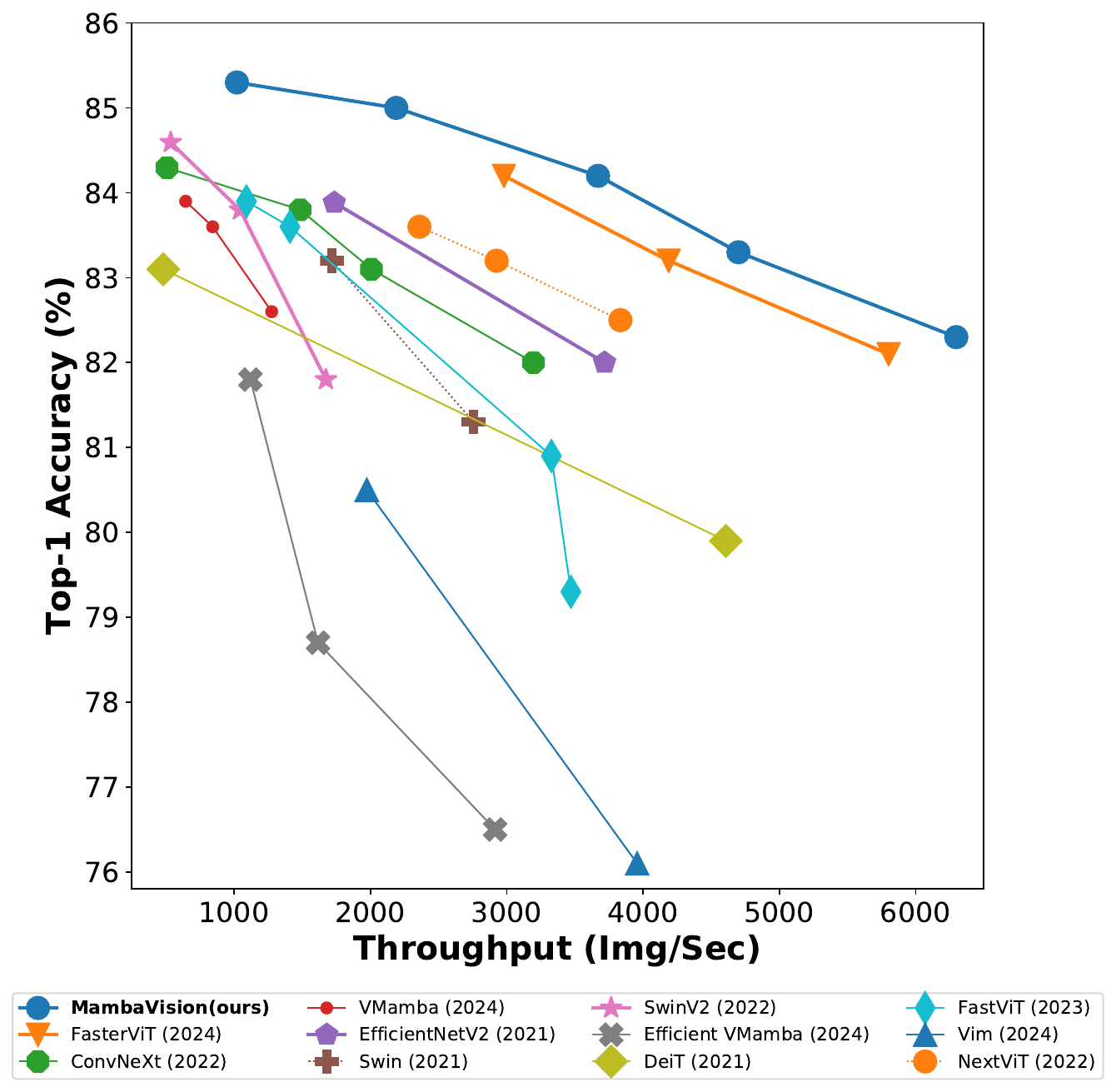}
    \put(70.6, 70.4){\smallnamefont{\textbf{MambaVision-T}}} 
   \put(65, 77.4){\smallnamefont{\textbf{MambaVision-S}}}
   \put(52.6, 84.2){\smallnamefont{\textbf{MambaVision-B}}}
   \put(34.6, 90.1){\smallnamefont{\textbf{MambaVision-L}}}
   \put(12.6, 92.6){\smallnamefont{\textbf{MambaVision-L2}}}
   \put(56.1, 22.2){\smallnamefont{Vim-T}}
   \put(32.0, 55.8){\smallnamefont{Vim-S}}
   \put(32.1, 20.2){\smallnamefont{Efficient VMamba-T}}
   \put(18.1, 36.2){\smallnamefont{Efficient VMamba-S}}
   \put(12.2, 60.2){\smallnamefont{Efficient VMamba-B}}
   \put(18.2, 67.2){\smallnamefont{VMamba-T}}
   \put(14.2, 75.2){\smallnamefont{VMamba-S}}
   \put(45.1, 41.2){\smallnamefont{FastViT-T12}}
   \put(49.1, 59.0){\smallnamefont{FastViT-SA12}}
   \put(54.1, 70.7){\smallnamefont{NextViT-S}}
   \put(44.1, 75.9){\smallnamefont{NextViT-B}}
   \put(37.1, 78.9){\smallnamefont{NextViT-L}}
   \put(60.4, 45.7){\smallnamefont{DeiT-S}}
   \put(12.2, 70.2){\smallnamefont{DeiT-B}}

     \tiny
  \end{overpic}
    \end{minipage}\hfill
\caption{\textbf{Top-1 accuracy \vs image throughput comparisons on ImageNet-1K dataset.} The MambaVision models achieve a new Pareto front for Top-1 accuracy and image throughput tradeoff. Specifically, MambaVision variants outperform Mamba-based models such as VMamba and Vim, sometimes by a significant margin. For all models, image throughput is measured on an A100 NVIDIA GPU with a batch size of 128.}
\label{fig:fig1}
\end{figure}

\section{Introduction}
In recent years, Transformers~\cite{vaswani2017attention} have become the de facto architecture across different domains including computer vision, natural language processing, speech processing, and robotics. The versatility of the Transformer architecture, primarily due to its attention mechanism and flexibility, makes it highly suitable for multimodal learning tasks, where integrating and processing information from various modalities is essential. Despite these benefits, the quadratic complexity of the attention mechanism with respect to sequence length makes Transformers computationally expensive to train and deploy. Recently, Mamba~\cite{gu2023mamba} proposed a new State Space Model (SSM) that achieves linear time complexity and outperforms or matches Transformers in different language modeling tasks~\cite{gu2023mamba}. The core contribution of Mamba is a novel selection mechanism that enables efficient input-dependent processing of long sequences with hardware-aware considerations. Recently, a number of Mamba-based backbones~\cite{zhu2024vision, pei2024efficientvmamba} have been proposed to leverage the strengths of its SSM formulation in vision tasks such as image classification and semantic segmentation. However, the Mamba's autoregressive formulation, while effective for tasks requiring sequential data processing, faces limitations in computer vision tasks that benefit from a full receptive field: (1) Unlike sequences where order matters, image pixels do not have a sequential dependency in the same way. Instead, spatial relationships are often local and need to be considered in a more parallel and integrated manner. Hence, this results in inefficiency for processing spatial data (2) An autoregressive model like Mamba processes data step-by-step, limiting its ability to capture and utilize global context in one forward pass. In contrast, vision tasks often require understanding the global context to make accurate predictions about local regions. Vision Mamba (Vim)~\cite{zhu2024vision} and others have proposed modifications such as bidirectional SSMs to address lack of global context and spatial understanding. While bidirectional SSMs have the potential to capture more comprehensive context, they introduce significant latency due to the need to process the entire sequence before making predictions. Additionally, the increased complexity can lead to challenges in training, risk of overfitting, and may not always result in better accuracy. Due to these pitfalls, backbones with Vision Transformer (ViT) and Convolutional Neural Network (CNN) architectures still outperform best Mamba-based vision models on different vision tasks.

In this work, we systematically re-design the Mamba block to make it more suitable for vision tasks. We propose a hybrid architecture that consists of our proposed formulation (\textit{i.e.} MambaVision Mixer and MLP) as well as Transformer blocks. Specifically, we study different integration patterns such as adding the Transformer blocks in an iso-parameter manner to earlier, middle, and final layers as well as every $l$ layers. Our analysis shows that leveraging several self-attention blocks at the final stages can significantly enhance the capability to capture global context and long-range spatial dependencies. As shown in Sec.~\ref{sec:results}, using a hybrid architecture also results in higher image throughput compared to both pure Mamba and ViT-based models. We introduce the MambaVision model which consists of a multi-resolution architecture and leverages CNN-based residual blocks for fast feature extraction of larger resolution features. As shown in Fig.~\ref{fig:fig1}, the MambaVision achieves a new SOTA Pareto front in terms of ImageNet-1K Top-1 accuracy and image throughput, outperforming Mamba, CNN, and ViT-based models, sometimes by a significant margin. In downstream tasks such as object detection, instance segmentation, and semantic segmentation, models with MambaVision backbones outperform comparably-sized counterparts on MS COCO and ADE20 datasets, respectively. Hence, it validates the effectiveness and versatility of MambaVision as an efficient backbone.

To the best of our knowledge, MambaVision is the \emph{first} effort to study and develop a hybrid architecture comprising of both Mamba and Transformers for computer vision applications. Our main contributions in this work are summarized as follows:

\begin{itemize}[noitemsep,nosep]
\item We introduce a redesigned vision-friendly Mamba block, improving accuracy and image throughput over the original Mamba architecture.

\item We present a systematic investigation of integration patterns for Mamba and Transformer blocks, and demonstrate that incorporating self-attention blocks at the final stages significantly improves the model's ability to capture global context and long-range spatial dependencies.

\item We introduce MambaVision, which is a novel hybrid Mamba-Transformer model. The hierarchical MambaVision achieves a new SOTA Pareto front on the ImageNet-1K dataset in terms of Top-1 and image throughput.

\end{itemize}

\begin{figure*}[t]
    \centering
    \includegraphics[width=0.99\linewidth]{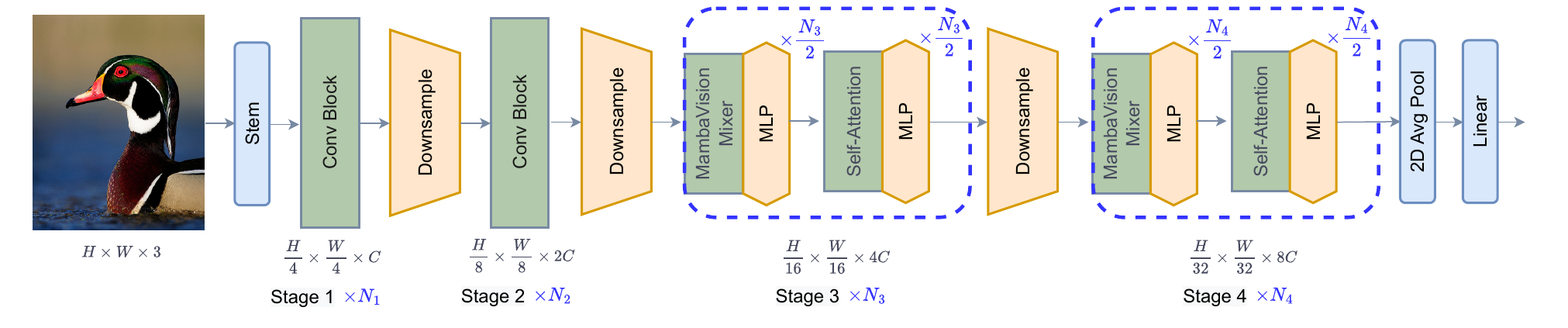}
    \caption{The architecture of hierarchical MambaVision models. The first two stages use residual convolutional blocks for fast feature extraction. Stages 3 and 4 employ both MambaVision and Transformer blocks. Specifically, given $N$ layers, we use $\frac{N}{2}$ MambaVision and MLP blocks, which are followed by additional $\frac{N}{2}$ Transformer and MLP blocks. The Transformer blocks in the final layers allow for recovering lost global context and capturing long-range spatial dependencies.
  }
    \label{fig:model_architecture}
\end{figure*}

\section{Related work}

\textbf{Conv-Based}.  CNNs have been the cornerstone of computer vision since the introduction of AlexNet~\cite{krizhevsky2012imagenet}. Recent efforts have focused on modernizing CNN architectures with Transformer-inspired principles. ConvNeXt~\cite{liu2022convnet} demonstrated competitive performance with Transformers by redesigning ResNet~\cite{he2016deep} with increased width, larger kernels, and layer normalization. RegNetY~\cite{radosavovic2020designing} introduced systematic network design through design space analysis, while EfficientNetV2~\cite{tan2021efficientnetv2} leveraged neural architecture search and progressive learning for better efficiency trade-offs. Despite strong performance, these CNN models inherently lack the global receptive field needed for capturing long-range dependencies.

\textbf{Transformer-Based}. ViTs~\cite{dosovitskiy2020image} marked a significant shift in computer vision by introducing self-attention mechanisms for enlarged receptive fields. However, ViTs initially lacked the inherent advantages of CNNs and required extensive training data. To address these limitations, DeiT~\cite{touvron2021training} introduced distillation-based training for improved accuracy on smaller datasets, while Swin Transformer~\cite{liu2021swin} proposed a hierarchical architecture using shifted windows for self-attention, effectively balancing local and global context. Models such as Twins~\cite{chu2021twins} and PVT~\cite{wang2021pyramid} further enhanced efficiency through spatially separable self-attention and hierarchical structures with patch embedding. Despite these advances, the quadratic complexity of self-attention operations in these models continued to pose efficiency challenges.

\textbf{Conv-Transformer}. The complementary strengths of CNNs and ViTs inspired hybrid architectures. CoAT~\cite{xu2021co} and CrossViT~\cite{chen2021crossvit} demonstrated enhanced feature learning by combining convolutions with self-attention, while NextViT~\cite{li2022next} systematically incorporated CNN-like processing into Transformers. Recent efforts like EfficientFormer~\cite{li2022efficientformer} and FasterViT~\cite{hatamizadeh2023fastervit} focused on optimizing efficiency-accuracy trade-offs, achieving competitive performance with high throughput through carefully designed hybrid architectures.

\textbf{Mamba-Based}. Since the introduction of Mamba, a number of efforts have been proposed to leverage its capability for vision applications. Vim~\cite{zhu2024vision} introduced a bidirectional SSM formulation that processes tokens in both forward and backward directions to capture global context and improve spatial understanding. However, this bidirectional approach faces significant limitations: it increases computational overhead, slows down training and inference times, and struggles to effectively combine information from multiple directions without losing global context. In contrast, MambaVision achieves superior results using a single forward pass with our redesigned Mamba block, demonstrating significantly better ImageNet Top-1 accuracy and throughput.
VMamba~\cite{liu2024vmamba} proposed a generic Mamba-based vision backbone featuring a Cross-Scan Module (CSM). This module implements a four-way selective scan methodology (from upper-left and lower-right to opposite directions) to integrate information from surrounding tokens and capture global context. VMamba also incorporates architectural modifications like depth-wise convolutions and a hierarchical multi-resolution structure. While the CSM module is designed for vision tasks, its receptive field remains constrained by the cross-scan paths. MambaVision offers several advantages over VMamba: our mixer design is simpler yet captures both short and long-range dependencies more effectively, we employ CNN-based layers for fast feature extraction rather than using uniform block structures across all stages, and we achieve superior performance with significantly higher throughput.
EfficientVMamba~\cite{pei2024efficientvmamba} uses SSMs for larger resolutions and CNNs for lower ones, while MambaVision takes the opposite approach with CNNs at higher resolutions and SSM/self-attention at lower ones, leading to significantly better accuracy and throughput. Similarly, while SiMBA~\cite{patro2024simba} addresses Mamba's stability through EinFFT channel modeling, it doesn't fully address spatial understanding limitations, resulting in lower performance compared to MambaVision's comprehensive design.

\section{Methodology}
\label{sec:method}
\subsection{Macro Architecture}
In this section, we introduce MambaVision which is our proposed novel architecture with SOTA performance on ImageNet-1K dataset. As illustrated in Fig.~\ref{fig:model_architecture}, MambaVision has a hierarchical architecture consisting of $4$ different stages. The first two stages consist of CNN-based layers for fast feature extraction at higher input resolutions, while stage $3$ and $4$ include the proposed MambaVision and Transformer blocks. Specifically, given an image of size $H \times W \times 3$, the input is first converted into overlapping patches with size $\frac{H}{4} \times \frac{W}{4} \times C$ and projected into a $C$ dimensional embedding space by the stem which consists of two consecutive $3 \times 3$ CNN layers with stride of $2$. The downsampler in between stages consists of a $3 \times 3$ CNN layer with stride $2$ which reduces the image resolution by half. Furthermore, the CNN blocks in stages $1$ and $2$ follow a generic residual block formulation according to the following 
\begin{align}
\begin{split}
\mathbf{\hat{z}} & = \text{GELU}(\text{BN}(\text{Conv}_{3\times3}(\mathbf{z}))), \\
\mathbf{z} & = \text{BN}(\text{Conv}_{3\times3}(\mathbf{\hat{z}})) + \mathbf{z},
\label{eq:fused_convmb}
\end{split}
\end{align}
GELU and BN denote Gaussian Error Linear Unit activation function~\cite{hendrycks2016gaussian} and batch normalization~\cite{ioffe2015batch}, respectively. Please see the supplementary materials for further details regarding MambaVision macro architecture. 

\begin{algorithm*}[t]
\caption{PyTorch-like pseudo-code for MambaVision mixer}
\label{alg:mamba_vision_mixer}
\definecolor{codeblue}{rgb}{0.25,0.5,0.5}
\definecolor{codekw}{rgb}{0.85, 0.18, 0.50}
\lstset{
  backgroundcolor=\color{white},
  basicstyle=\fontsize{7.5pt}{7.5pt}\ttfamily\selectfont,
  columns=fullflexible,
  breaklines=true,
  captionpos=b,
  commentstyle=\fontsize{7.5pt}{7.5pt}\color{codeblue},
  keywordstyle=\fontsize{7.5pt}{7.5pt}\color{codekw},
}
\begin{lstlisting}[language=python]
import torch
import math
import torch.nn as nn
import torch.nn.functional as F
from einops import rearrange, repeat

class MambaVisionMixer(nn.Module):
    def __init__(self, dim, d_state=16, kernel_size=3):
        super().__init__()
        self.d_state = d_state
        self.dt_rank = math.ceil(dim / 16)
        self.in_proj = nn.Linear(self.d_model, self.d_inner)
        self.x_proj = nn.Linear(dim//2, self.dt_rank + self.d_state *2)
        self.conv1d_x = nn.Conv1d(dim//2, dim//2, kernel_size=kernel_size, padding='same', groups=dim//2)
        self.conv1d_z = nn.Conv1d(dim//2, dim//2, kernel_size=kernel_size, padding='same', groups=dim//2)
        self.dt_proj = nn.Linear(self.dt_rank, dim//2)
        dt = torch.exp(torch.rand(self.dim//2) * (math.log(dt_max) - math.log(dt_min)) + math.log(dt_min))
        A_log = torch.log(repeat(torch.arange(1, self.d_state + 1), n -> d n, d=dim//2))
        self.A_log = nn.Parameter(A_log)
        self.D = nn.Parameter(torch.ones(dim//2))
        self.out_proj = nn.Linear(dim, dim)

    def forward(self, hidden_states):
        xz = rearrange(self.in_proj(hidden_states), b l d -> b d l)
        x, z = xz.chunk(2, dim=1)
        A = -torch.exp(self.A_log)
        x = F.silu(self.conv1d_x(x))
        z = F.silu(self.conv1d_z(z))
        seqlen = hidden_states.shape[1]
        x_dbl = self.x_proj(rearrange(x, b d l -> (b l) d))
        dt, B, C = torch.split(x_dbl, [self.dt_rank, self.d_state, self.d_state], dim=-1)
        dt = rearrange(self.dt_proj(dt), (b l) d -> b d l, l=seqlen)
        B = rearrange(B, (b l) dstate -> b dstate l, l=seqlen)
        C = rearrange(C, (b l) dstate -> b dstate l, l=seqlen)
        x_ssm = selective_scan_fn(x, dt, A, B, C, D)
        hidden_states = rearrange(torch.cat([x_ssm, z], dim=1), b d l -> b l d)
        return self.out_proj(hidden_states)
\end{lstlisting}
\end{algorithm*}

\subsection{Micro Architecture}
In this section, we first revisit the preliminaries of Mamba and SSMs. We then present the micro design of the architecture in stages 3 and 4 and discuss MambaVision formulation in more details.   
\subsubsection{Mamba Preliminaries}
In Mamba, a 1D continuous input $x(t) \in \mathbb{R}$ is transformed into $y(t) \in \mathbb{R}$ via a learnable hidden state $h(t) \in \mathbb{R}^{M}$ with parameters $\bm{A}\in\mathbb{R}^{M\times M}$, $\bm{B}\in\mathbb{R}^{M\times 1}$ and $\bm{C}\in\mathbb{R}^{1 \times M}$ according to
\begin{align}\label{eq:ode}
    \begin{split}
        h'(t) &= \bm{A}h(t) + \bm{B}x(t),\\
        y(t) &= \bm{C}h(t), 
    \end{split}
\end{align}
\paragraph{Discretization} The continuous parameters $\bm{A}$, $\bm{B}$ and $\bm{C}$ in the above formulation are further converted into discrete parameters for better computational efficiency~\cite{gu2021combining}. Specifically, assuming a timescale $\Delta$, a zero-order hold rule can be applied to obtain discrete parameters $\bm{\bar{A}}\in\mathbb{R}^{M\times M}$, $\bm{\bar{B}} \in \mathbb{R}^{M\times 1}$ and $\bm{\bar{C}}\in\mathbb{R}^{1 \times M}$ according to
\begin{equation} \label{eq:discretization}
\begin{aligned}
&\bar{\bm{A}} = \exp{(\Delta \bm{A})}, \\
&\bar{\bm{B}} =(\Delta \bm{A})^{-1}(\exp{(\Delta \bm{A})} - \bm{I}) \cdot (\Delta \bm{B}), \\
&\bar{\bm{C}} = \bm{C}, \\
\end{aligned}
\end{equation}
The Eq.~\ref{eq:ode} can then be expressed with discrete parameters as
\begin{equation} \label{eq:discretization2}
\begin{aligned}
&h(t) = \bar{\bm{A}}h(t-1) + \bar{\bm{B}}x(t), \\
&y(t) = \bar{\bm{C}}h(t),
\end{aligned}
\end{equation}
In addition, for an input sequence with size $T$, a global convolution with kernel $\bm{\overline{K}}$ can be applied for computing the output of Eq.~\ref{eq:discretization2} as in the following
\begin{align}
    \begin{split}
       \bm{\overline{K}} &= (\bm{C}\bm{\overline{B}},\bm{C}\overline{\bm{A}\bm{B}}, ..., \bm{C}\bm{\overline{A}}^{T-1}\bm{\overline{B}}), \\
        \quad \bm{y} &= \bm{x} * \bm{\overline{K}}, \\
    \end{split}
\end{align}
\paragraph{Selectivity} Mamba further extends the S4 formulation by introducing a selection mechanism which allows for input-dependant sequence processing. This allows the model's parameters $\bm{B}$, $\bm{C}$ and $\Delta$ to be adjusted dynamically according to the inputs and filter out irrelevant information. Further discretization details are provided in~\cite{gu2023mamba}.

\subsubsection{Layer Architecture}
Assuming an input $X \in \mathbb{R}^{T \times C}$ with sequence length $T$ with embedding dimension $C$, the output of layer $n$ in stages 3 and 4 can be computed as in
\begin{equation}
\begin{array}{l}
\hat{{X}}^{n}=\text{Mixer}(\text{Norm}({X}^{n-1}))+{X}^{n-1}, \\
{X}^{n}=\text{MLP}(\text{Norm}(\hat{{X}}^{n}))+\hat{{X}}^{n}, \\
\end{array}
\label{eq:eq_layer}
\end{equation}
Norm and Mixer denote the choices of layer normalization and token mixing blocks, respectively. Without loss of generality, Layer Normalization is used for Norm. Given $N$ layers, the first $\frac{N}{2}$ layers employ MambaVision mixer blocks while the remaining $\frac{N}{2}$ layers employ self-attention. We describe the details of each mixer in the following.

\begin{figure}[t]
    \centering
    \includegraphics[width=0.7\linewidth]{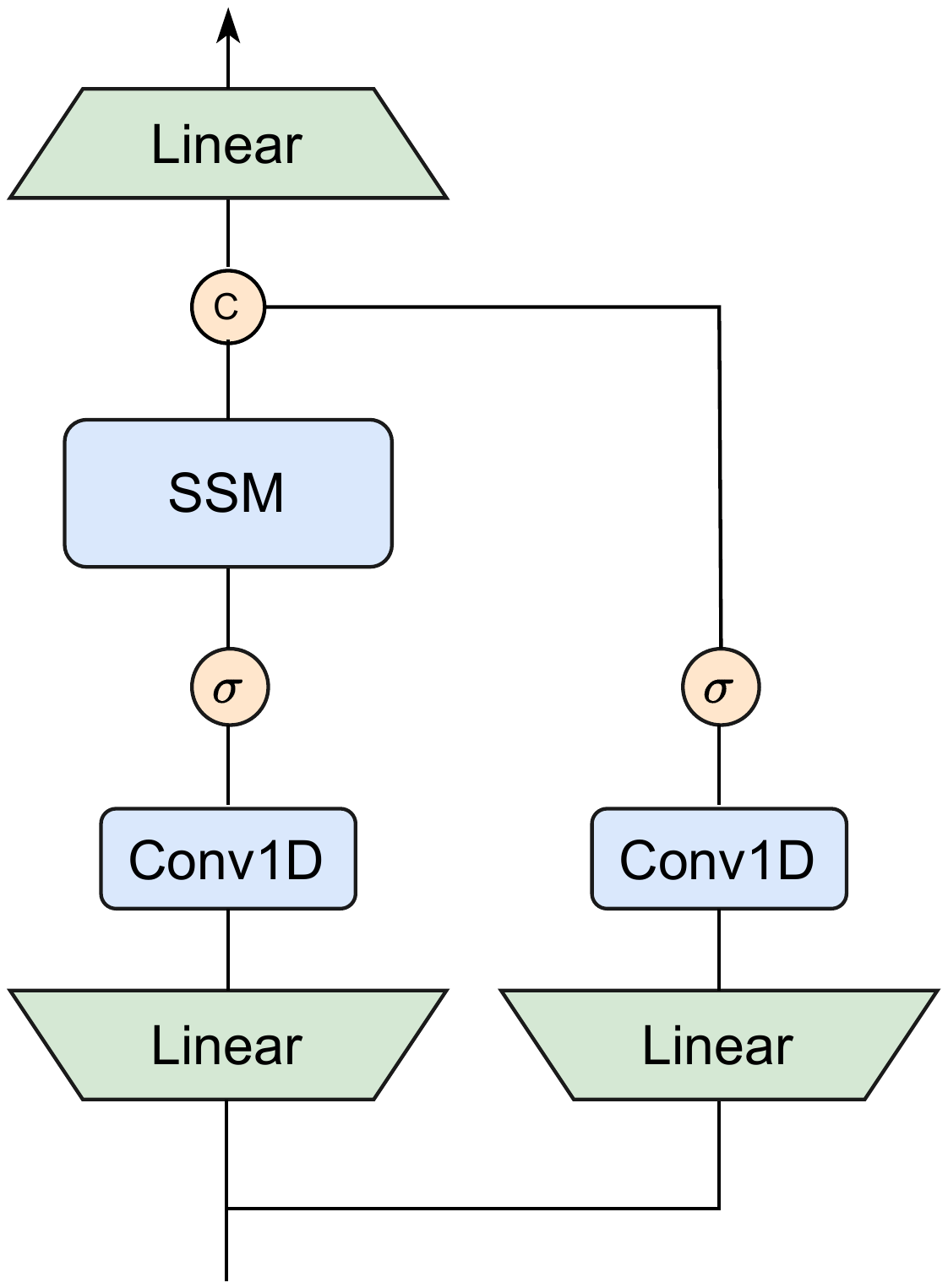}
    \caption{
    \textbf{Architecture of MambaVision block.} In addition to replacing causal Conv layer with their regular counterparts, we create a symmetric path without SSM as a token mixer to enhance the modeling of global context.    
    }
    \label{fig:mamba_vision}
\end{figure}

\paragraph{MambaVision Mixer} As shown in Fig.~\ref{fig:mamba_vision}, we redesigned the original Mamba mixer to make it more suitable for vision tasks. First, we propose to replace the causal convolution with regular convolution, since it limits the influence to one direction, which is unnecessary and restrictive for vision tasks. In addition, we added a symmetric branch without SSM, consisting of an additional convolution and Sigmoid Linear Unit (SiLU)~\cite{elfwing2018sigmoid} activation, to compensate for any content lost due to the sequential constraints of SSMs. We then concatenate the output of both branches and project it via a final linear layer. This combination ensures that the final feature representation incorporates both the sequential and spatial information, leveraging the strengths of both branches. We note that the output of each branch is projected into an embedding space with size $\frac{C}{2}$ (\textit{i.e.} half the size of original embedding dimension) to maintain similar number of parameters to the original block design. Given an input $X_{in}$, the output of MambaVision mixer $X_{out}$ is computed according to
\begin{equation}
\begin{array}{l}
{X_{1}}=\text{Scan}(\sigma(\text{Conv}(\text{Linear}(C, \frac{C}{2})({X_{in}})))), \\
{X_{2}}=\sigma(\text{Conv}(\text{Linear}(C, \frac{C}{2})({X_{in}}))), \\
{X_{out}}=\text{Linear}(\frac{C}{2},C)(\text{Concat}(X_{1}, X_{2})),
\end{array}
\label{eq:eq_mixer1}
\end{equation}
$\text{Linear}(C_{in}, C_{out})(\cdot)$ denotes a linear layer with $C_{in}$ and $C_{out}$ as input and output embedding dimensions, $\text{Scan}$ is the selective scan operation as in \cite{gu2023mamba} and $\sigma$ is the activation function for which SiLU is used. In addition, \text{Conv} and \text{Concat} represent 1D convolution and concatenation operations. In Algorithm~\ref{alg:mamba_vision_mixer}, we present a PyTorch-like pseudo-code for MambaVision mixer. In general, our proposed modification leads to richer feature representations, better generalization, and improved performance on computer vision tasks. We have also experimentally validated the effectiveness of each of our design choices in Sec.~\ref{sec:abl}.   

\paragraph{Self-attention} We use a generic multihead self-attention mechanism in accordance to
\begin{equation}
    {\rm Attention}({Q}, {K}, {V}) = {\rm Softmax}(\frac{{Q}{K}^\mathsf{T}}{\sqrt{d_{h}}}){V}.
    \label{eqn:mhsa}
\end{equation}
$Q, K, V$ denote query, key and value respectively and $d_{h}$ is the number of attention heads. In addition, our framework allows for computing the attention in a windowed manner similar to previous efforts~\cite{liu2021swin, liu2022swin} (see the supplementary materials for window size ablation study).

\section{Experiments}
Image classification experiments are conducted on the ImageNet-1K dataset~\cite{deng2009imagenet}. We followed the standard training recipe of previous efforts~\cite{liu2021swin,yang2021focal,hatamizadeh2023global} to allow for a comparable analysis of performance across different models. Specifically, all models have been trained for 300 epochs using 32 A100 GPUs. The self-attention formulation in stages 3 and 4 of all MambaVision variants use a window size of 14 and 7, respectively. For detailed training configurations, please see the provided anonymous code repository. To evaluate the performance of downstream tasks, we used our pre-trained models as backbones for object detection, instance segmentation, and semantic segmentation tasks using the MS COCO dataset~\cite{lin2014microsoft} and ADE20K dataset~\cite{zhou2017scene}, respectively. Specifically, for object detection and instance segmentation, we used Cascade Mask-RCNN~\cite{he2017mask} head with hyperparameters such as ×3 LR schedule. For semantic segmentation, we used a UperNet network~\cite{xiao2018unified} head and 8 A100 GPUs for all experiments.

\begin{table}[t]
\renewcommand\arraystretch{.9}
\centering
\tablestyle{6pt}{1.00}
\caption{Comparison of classification benchmarks on \textbf{ImageNet-1K} dataset~\cite{deng2009imagenet}. Image throughput is measured on A100 GPU with a batch size of 128.} 
\resizebox{0.96\linewidth}{!}{
\setlength{\tabcolsep}{.5mm}{
\begin{tabular}[t]{lccccc}
\toprule
Model &  Image Size& \#Params & FLOPs & Throughput& Top-1 \\
      &  (Px) & (M) & (G) & (Img/Sec) & (\%) \\
\midrule
\multicolumn{6}{c}{Conv-Based} \\
\midrule
ConvNeXt-T~\cite{liu2022convnet} & 224 &\ \  28.6 &\ \ 4.5 & 3196 & 82.0\\
ConvNeXt-S~\cite{liu2022convnet} & 224 &\ \  50.2 &\ \ 8.7 & 2008 & 83.1\\
ConvNeXt-B~\cite{liu2022convnet} & 224 &\ \  88.6 & 15.4 & 1485 & 83.8\\
RegNetY-040~\cite{radosavovic2020designing} & 288 &\ \  20.6 &\ \  6.6 & 3227 & 83.0\\
ResNetV2-101~\cite{wightman2021resnet}& 224 &\ \  44.5 &\ \  7.8 & 4019 & 82.0\\
EfficientNetV2-S~\cite{tan2021efficientnetv2} & 384 &\ \  21.5 &\ \  8.0 & 1735 & 83.9\\
\midrule
\multicolumn{6}{c}{Transformer-Based} \\
\midrule
Swin-T~\cite{liu2021swin} & 224 &\ \  28.3 &\ \  4.4 & 2758 & 81.3\\
Swin-S~\cite{liu2021swin} & 224 &\ \  49.6 &\ \  8.5 & 1720 & 83.2\\
SwinV2-T~\cite{liu2022swin} & 256 &\ \  28.3 &\ \  4.4 & 1674 & 81.8\\
SwinV2-S~\cite{liu2022swin} & 256 &\ \  49.7 &\ \  8.5 & 1043 & 83.8\\
SwinV2-B~\cite{liu2022swin} & 256 &\ \  87.9 & 15.1 &\ \  535 & 84.6\\
TNT-S~\cite{han2021transformer} & 224 &\ \  23.8 &\ \  4.8 & 1478 & 81.5\\
Twins-S~\cite{chu2021twins} & 224 &\ \  24.1 &\ \  2.8 & 3596 & 81.7\\
Twins-B~\cite{chu2021twins} & 224 &\ \  56.1 &\ \  8.3 & 1926 & 83.1\\
Twins-L~\cite{chu2021twins} & 224 &\ \  99.3 & 14.8 & 1439 & 83.7\\
DeiT-B~\cite{touvron2021training} & 224 &\ \  86.6 & 16.9 & 2035 & 82.0\\
DeiT3-L~\cite{touvron2022deit} & 224 & 304.4 & 59.7 &\ \ 535  & 84.8	\\
PoolFormer-M58~\cite{yu2022metaformer} & 224 &\ \  73.5 & 11.6 &\ \  884 & 82.4\\
\midrule
\multicolumn{6}{c}{Conv-Transformer} \\
\midrule
CoaT-Lite-S~\cite{xu2021co} & 224 &\ \  19.8 & \ \  4.1 & 2269 & 82.3\\
CrossViT-S~\cite{chen2021crossvit} & 240 & 26.9 & 5.1 & 2832 & 81.0\\
CrossViT-B~\cite{chen2021crossvit} & 240 & 105.0 & 20.1 & 1321 & 82.2\\
Visformer-S~\cite{chen2021visformer} & 224 &\ \  40.2 &\ \  4.8 & 3676 & 82.1\\
NextViT-S~\cite{li2022next} & 224 & 31.7 & 5.8 & 3834 & 82.5\\
NextViT-B~\cite{li2022next}  & 224 &\ \  44.8 & \ \ 8.3 & 2926 & 83.2\\
NextViT-L~\cite{li2022next} & 224 &\ \  57.8 & 10.8 & 2360 & 83.6\\
EfficientFormer-L1~\cite{li2022efficientformer} & 224 & 12.3 & 1.31 & 6220 & 79.2\\
EfficientFormer-L3~\cite{li2022efficientformer} & 224 &\ \  31.4 & \ \ 3.9 & 2845 & 82.4\\
EfficientFormer-L7~\cite{li2022efficientformer} & 224 &\ \  82.2 & 10.2 & 1359 & 83.4\\
MaxViT-B~\cite{tu2022maxvit} & 224 &\ \  120.0 & \ \ 23.4 & 507 & 84.9\\
MaxViT-L~\cite{tu2022maxvit} & 224 &\ \  212.0 & \ \ 43.9 & 376 & 85.1\\
FasterViT-1~\cite{hatamizadeh2023fastervit} &  224 &\ \  53.4 &\ \  5.3 & 4188 &  83.2\\
FasterViT-2~\cite{hatamizadeh2023fastervit} &  224 &\ \  75.9 &\ \  8.7 & 3161 &  84.2 \\
FasterViT-3~\cite{hatamizadeh2023fastervit} &  224 & 159.5 & 18.2 & 1780 &  84.9 \\

\midrule
\multicolumn{6}{c}{Mamba-Based} \\
\midrule
Vim-T~\cite{zhu2024vision} & 224 & 7.0 & - & 3957 & 76.1\\
Vim-S~\cite{zhu2024vision} & 224 & 26.0 & - & 1974 & 80.5\\
EfficientVMamba-T~\cite{pei2024efficientvmamba} & 224 &6.0 & 0.8 & 2904 & 76.5\\
EfficientVMamba-S~\cite{pei2024efficientvmamba} & 224 &11.0 & 1.3 & 1610 & 78.7\\
EfficientVMamba-B~\cite{pei2024efficientvmamba} & 224 &33.0 & 4.0 & 1482 & 81.8\\
SiMBA-S~\cite{patro2024simba} & 224 & 15.3 & 2.4 & 826 & 81.7\\
SiMBA-B~\cite{patro2024simba} & 224 & 22.8 & 4.2 & 624 & 83.5\\
VMamba-T~\cite{liu2024vmamba} & 224 &30.0 & 4.9 & 1282 & 82.6\\
VMamba-S~\cite{liu2024vmamba} & 224 &50.0 & 8.7 & 843 & 83.6\\
VMamba-B~\cite{liu2024vmamba} & 224 &89.0 & 15.4 & 645 & 83.9\\
\midrule
\multicolumn{6}{c}{\textbf{MambaVision}} \\
\midrule
\rowcolor{Gray} 
\textbf{MambaVision-T} & 224 & \ \ 31.8 &\ \  4.4 & \textbf{6298} &  \textbf{82.3} \\
\rowcolor{Gray} 
\textbf{MambaVision-T2} &  224 &\ \  35.1 &\ \  5.1 & \textbf{5990} &  \textbf{82.7}\\
\rowcolor{Gray} 
\textbf{MambaVision-S} &  224 &\ \  50.1 &\ \  7.5 & \textbf{4700} &  \textbf{83.3} \\
\rowcolor{Gray}
\textbf{MambaVision-B} &  224 & 97.7 & 15.0 & \textbf{3670} &  \textbf{84.2} \\
\rowcolor{Gray}
\textbf{MambaVision-L} &  224 & 227.9 & 34.9 &\ \  \textbf{2190} &  \textbf{85.0} \\
\rowcolor{Gray}
\textbf{MambaVision-L2} &  224 & 241.5 & 37.5 &\ \  \textbf{1021} &  \textbf{85.3} \\

\bottomrule
\end{tabular}}
}
\label{tab:imgnet}
\end{table}

\setlength{\tabcolsep}{8pt}
\begin{table*}[!t]
    \centering
    \resizebox{0.8\linewidth}{!}{
    \begin{tabular}{l|cc|cccccc}
        \toprule
        Backbone & Params (M) & FLOPs (G) & $\text{AP}^{\text{box}}$ & $\text{AP}^{\text{box}}_{50}$ & $\text{AP}^{\text{box}}_{75}$ & $\text{AP}^{\text{mask}}$ & $\text{AP}^{\text{mask}}_{\text{50}}$ & $\text{AP}^{\text{mask}}_{\text{75}}$ \\ 
        \midrule
        DeiT-Small/16~\cite{touvron2021training} & 80 & 889 & 48.0 & 67.2 & 51.7 & 41.4 & 64.2 & 44.3 \\
		ResNet-50~\cite{he2016deep} & 82 & 739 & 46.3 & 64.3 & 50.5 & 40.1 & 61.7 & 43.4 \\
        Swin-T~\cite{liu2021swin} & 86 & 745 & 50.4 & 69.2 & 54.7 & 43.7 & 66.6 & 47.3 \\
        ConvNeXt-T~\cite{liu2022convnet} & 86 & 741 & 50.4 & 69.1 & 54.8 & 43.7 & 66.5 & 47.3
        \\ \rowcolor{Gray}
        \textbf{MambaVision-T} & 86 & 740 & \textbf{51.1} & 70.0 & 55.6 & \textbf{44.3} & 67.3 & 47.9 \\
		\midrule
		X101-32~\cite{xie2017aggregated} & 101 & 819 & 48.1 & 66.5 & 52.4 & 41.6 & 63.9 & 45.2 \\
        Swin-S~\cite{liu2021swin} & 107 & 838  & 51.9 & 70.7 & 56.3 & 45.0 & 68.2 & 48.8 \\
        ConvNeXt-S~\cite{liu2022convnet} & 108 & 827 & 51.9 & 70.8 & 56.5 & 45.0 & 68.4 & 49.1
        \\ \rowcolor{Gray}
        \textbf{MambaVision-S} & 108 & 828 & \textbf{52.3} & 71.1 & 56.7 & \textbf{45.2} & 68.5 & 48.9 \\
        \midrule
        X101-64~\cite{xie2017aggregated} & 140 & 972 & 48.3 & 66.4 & 52.3 & 41.7 & 64.0 & 45.1 \\
        Swin-B~\cite{liu2021swin} & 145 & 982  & 51.9 & 70.5 & 56.4 & 45.0 & 68.1 & 48.9
        \\ 
        ConvNeXt-B~\cite{liu2022convnet} & 146 & 964 & 52.7 & 71.3 & 57.2 & 45.6 & 68.9 & 49.5 \\\rowcolor{Gray}
        \textbf{MambaVision-B} & 145 & 964 & \textbf{52.8} & 71.3 & 57.2 & \textbf{45.7} & 68.7 & 49.4 \\
        \bottomrule
    \end{tabular}
    }
    \caption{Object detection and instance segmentation benchmarks using Cascade Mask R-CNN~\cite{he2017mask} on \textbf{MS COCO} dataset~\cite{lin2014microsoft}. All models are trained by using a $3\times$ schedule and a crop resolution of $1280\times 800$.}
    \label{tab:cascademaskrcnn}
\end{table*}

\section{Results}
\label{sec:results}
\subsection{Image classification}
In Table~\ref{tab:imgnet}, we present the ImageNet-1K classification results. Specifically, we compare against different families of models such as Conv-based, Transformer-based, Conv-Transformer, and Mamba-based architectures and demonstrate that our model outperforms the previous efforts by a large margin, considering ImageNet Top-1 accuracy and image throughput. For example, MambaVision-B achieves higher accuracy (84.2\%) compared to ConvNeXt-B (83.8\%) and Swin-B (83.5\%), while also having significantly better image throughput. We observe similar trends in comparison to Mamba-based models. Specifically, MambaVision-B (84.2\%) outperforms VMamba-B (83.9\%) despite having considerably higher image throughput. We observe similar trends in performance comparisons with respect to other Mamba-based models. In addition, we would also like to note that although our main design goal has been to optimize the accuracy and throughput tradeoff, the MambaVision model variants have much lower FLOPs when compared to similarly-sized counterparts. For instance, MambaVision-B has 56\% less GFLOPs than MaxViT-B.

\subsection{Object Detection and Segmentation}
We evaluate our model's object detection and instance segmentation performance on the MS COCO dataset~\cite{lin2014microsoft}, as shown in Table~\ref{tab:cascademaskrcnn}. To comprehensively validate MambaVision's effectiveness, we trained models of varying sizes and compared them with popular vision backbones of comparable scale under identical conditions. Using a Cascade Mask R-CNN~\cite{he2017mask} head, all variants of MambaVision demonstrated superior performance compared to their counterparts. Specifically, MambaVision models outperform ConvNeXt-T by +0.7 and +0.6, ConvNeXt-S by +0.4 and +0.2 and ConvNeXt-B by +0.1 and +0.1 in terms of box Average Precision (AP) and mask AP, respectively. Similarly, MambaVision outperforms Swin-T by +0.7 and +0.6, Swin-S by +0.4 and +0.2 and Swin-B by +0.9 and +0.7 in terms of box AP and mask AP, respectively. For semantic segmentation, we evaluated the performance on the ADE20K dataset~\cite{zhou2017scene} using UPerNet~\cite{xiao2018unified}, as shown in Table~\ref{tab:ade_segmentation}. We observe that MambaVision models outperform similarly-sized competing models for different variants. For instance, MambaVision-T, MambaVision-S, and MambaVision-B outperform Swin-T, Swin-S, and Swin-B by +1.5, +0.6, and +1.0 in terms of mIoU, respectively. Notably, these improvements were achieved without extensive hyperparameter optimization for downstream tasks, highlighting MambaVision's potential as a robust backbone for various vision tasks, particularly in high-resolution scenarios. Moreover, our approach consistently attains higher mIoU than Focal Transformers across all scales while having comparable model sizes. 

\begin{table}
\centering
\resizebox{.8\linewidth}{!}{
\footnotesize
\setlength{\tabcolsep}{2.5pt}
  \begin{tabular}{lccc}
    \toprule
    Backbone  & Param (M) & FLOPs (G) & mIoU \\
    \midrule	 
    DeiT-Small/16~\citep{touvron2021training}  & 52 & 1099 & 44.0\\
    Swin-T~\citep{liu2021swin}  & 60 & 945 & 44.5\\
    ResNet-101~\citep{he2016deep}  & 86 & 1029 & 44.9\\
    Focal-T~\citep{yang2021focal}  & 62 & 998 & 45.8\\\rowcolor{Gray}
    \textbf{MambaVision-T}  & 55 & 945 & \textbf{46.0}\\
    \midrule
    Swin-S~\citep{liu2021swin}  & 81 & 1038 & 47.6\\
    Twins-SVT-B~\citep{chu2021twins}  & 89 & - & 47.7\\
    Focal-S~\citep{yang2021focal}  & 85 & 1130 & 48.0\\
    \rowcolor{Gray}
    \textbf{MambaVision-S}  & 84 & 1135 & \textbf{48.2}\\
    \midrule
    Swin-B~\citep{liu2021swin}  & 121 & 1188 & 48.1\\
    Twins-SVT-L~\citep{chu2021twins}  & 133 & - & 48.8\\
    Focal-B~\citep{yang2021focal}  & 126 & 1354 & 49.0\\
    \rowcolor{Gray}
    \textbf{MambaVision-B}  & 126 & 1342 & \textbf{49.1}\\
    \bottomrule

  \end{tabular} 
  }
    \caption{Semantic segmentation results with UperNet~\cite{xiao2018unified} model using \textbf{ADE20K} dataset. All models are trained using a crop resolution of $512\times 512$.}
    \label{tab:ade_segmentation}
\end{table}

\subsection{Ablation}
\label{sec:abl}
\begin{figure}[t]
    \centering
    \includegraphics[width=0.9\linewidth]{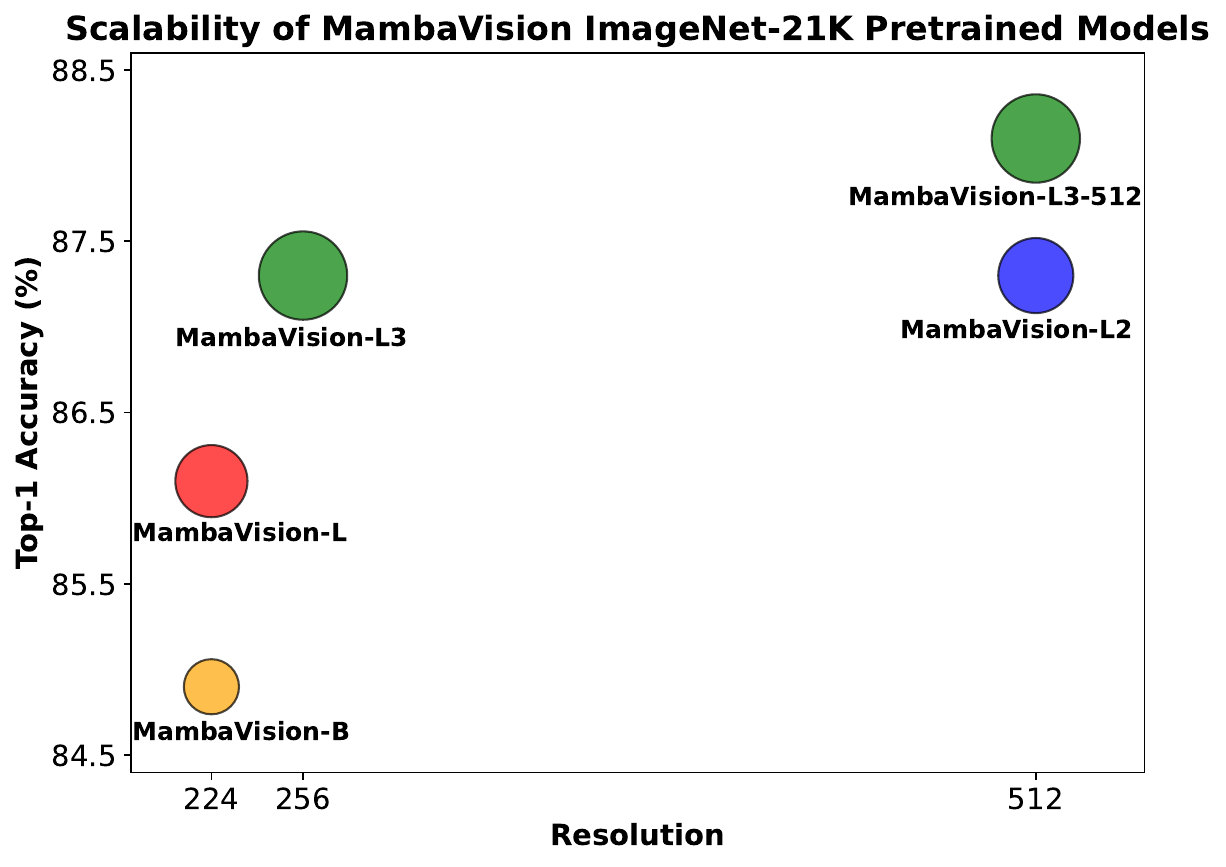}
    \caption{
    \textbf{Performance scalability} of MambaVision ImageNet-21K pretrained models with varying model sizes and resolutions.      
    }
    \label{fig:21k}
\end{figure}

\paragraph{Large-scale Training on ImageNet-21K} For the first time in any Mamba-based approach, our work (MambaVision) has scaled training to the large ImageNet-21K dataset with significantly bigger model sizes. As demonstrated in Fig.~\ref{fig:21k} the results are promising. Specifically, we observe meaningful improvements for the smaller MambaVision-B model (97.7M parameters), whose Top-1 accuracy increases from 84.2\% to 84.9\% at 224 resolution. In addition, pre-training and fine-tuning MambaVision-L raises its Top-1 accuracy from 85\% to 86.1\% at 224 resolution. We have also introduced a larger variant, MambaVision-L3 (739.6M parameters), which attains Top-1 accuracies of 87.3\% and 88.1\% at 256 and 512 resolutions, respectively. These results validate the scalability of our model across larger datasets, varying model sizes, and different image resolutions. To the best of our knowledge, MambaVision represents the first successful scaling of a Mamba-based vision architecture to ImageNet-21K with strong performance. This ability to scale is critical for real-world scenarios that rely on massive datasets, where bigger, more capable models are needed to achieve robust performance. We anticipate that MambaVision’s proven scalability will further encourage the adoption of Mamba-based models in industrial and large-scale research applications.

\paragraph{Design of Token Mixer} We conducted a comprehensive ablation study to systematically design the MambaVision token mixer. Our investigation focused on adapting the Mamba block for computer vision tasks, evaluating performance across classification, object detection, instance segmentation, and semantic segmentation. All experiments used a model architecture based on MambaVision-T configuration. As shown in Table~\ref{tab:abl_study1}, we began with the original Mamba formulation, which includes a causal convolution layer in the SSM branch (conv1) but lacks the additional convolution layer in our proposed symmetric branch (conv2). This baseline configuration achieved suboptimal performance across all metrics, with ImageNet Top-1 accuracy of 80.9\% (-1.8\%), MS COCO box AP of 44.8 (-1.6) and mask AP of 40.2 (-1.6), and ADE20K mIoU of 44.2\% (-1.4). We then replaced the causal convolution in the SSM branch (conv1) with a regular convolution layer, which improved performance across all metrics. Subsequently, we added conv2 layer while maintaining Mamba's original gating mechanism instead of concatenation, resulting in ImageNet Top-1 accuracy of 81.3\%, MS COCO box AP of 45.3 and mask AP of 41.0, and ADE20K mIoU of 45.7\%. Finally, implementing concatenation led to substantial improvements across all metrics, with gains of +1.0\% in ImageNet Top-1, +1.1 in box AP and +0.8 in mask AP for MS COCO, and +0.9 in mIoU for ADE20K. These results validate our hypothesis that concatenating outputs from both branches (SSM and non-SSM) enables the model to learn richer feature representations and enhance global context understanding. 

\paragraph{Hybrid Pattern} We conducted a comprehensive study examining various hybrid integration patterns between self-attention and MambaVision token mixers. All experiments maintained the MambaVision-T architecture layout with iso-parameter models for fair comparison, implementing hybrid functionality in stages 3 and 4. Initial experiments with a random integration pattern yielded suboptimal results with a Top-1 accuracy of 81.3\%, confirming our intuition that arbitrary self-attention placement may be ineffective. When we positioned self-attention blocks in the first N/2 layers of each stage (where N represents the total number of stage layers), performance improved by +0.2\% (81.5\%). A mixed layer pattern alternating between self-attention and MambaVision mixer blocks showed a slight performance decrease of -0.1\% (81.4\%), while reversing this order to MambaVision/self-attention improved accuracy to 81.6\%. Placing self-attention blocks in only the last N/4 layers of each stage yielded a significant improvement of +0.3\% (81.9\%), supporting our hypothesis that self-attention is most effective in the final layers. Further optimization revealed that extending self-attention to the last N/2 layers of each stage achieved the best performance at 82.3\%, indicating the importance of carefully balancing self-attention blocks with MambaVision layers for optimal representation learning.

\begin{table}
\small
    \caption{Systematic design of MambaVision token mixer. w/o and concat refer to "without" and concatenation. Conv1 and conv2 denote the conv operations in the SSM and additional symmetric branch as shown in Fig.~\ref{fig:mamba_vision}. COCO experiments are performed using Mask-RCNN~\cite{he2017mask} head and ×1 LR schedule.
    }
    \label{tab:abl_study1}
\centering
\resizebox{.99\linewidth}{!}{
\setlength{\tabcolsep}{2.5pt}
  \begin{tabular}{l|c|cc|c}
\Xhline{1.0pt}
 & \multicolumn{1}{c|}{ImageNet} & \multicolumn{2}{c|}{COCO} & \multicolumn{1}{c}{ADE20k} \\
 & top-1   & AP$^\text{box}$ & AP$^\text{mask}$ & mIoU \\
\hline
causal conv1 - w/o conv2  & 80.5 & 44.8 & 40.4 & 44.2 \\
conv1 - w/o conv2  & 80.9 & 45.0 & 40.8 & 44.7 \\
conv1 - conv2 - w/o concat  & 81.3 & 45.3 & 41.0 & 45.7 \\
\hline
\rowcolor{Gray}
conv1 - conv2 - concat & \textbf{82.3} & \textbf{46.4} & \textbf{41.8} & \textbf{46.0} \\

\Xhline{1.0pt}
\end{tabular}
  }
\end{table}

\begin{figure}[t]
    \centering
    \includegraphics[width=0.9\linewidth]{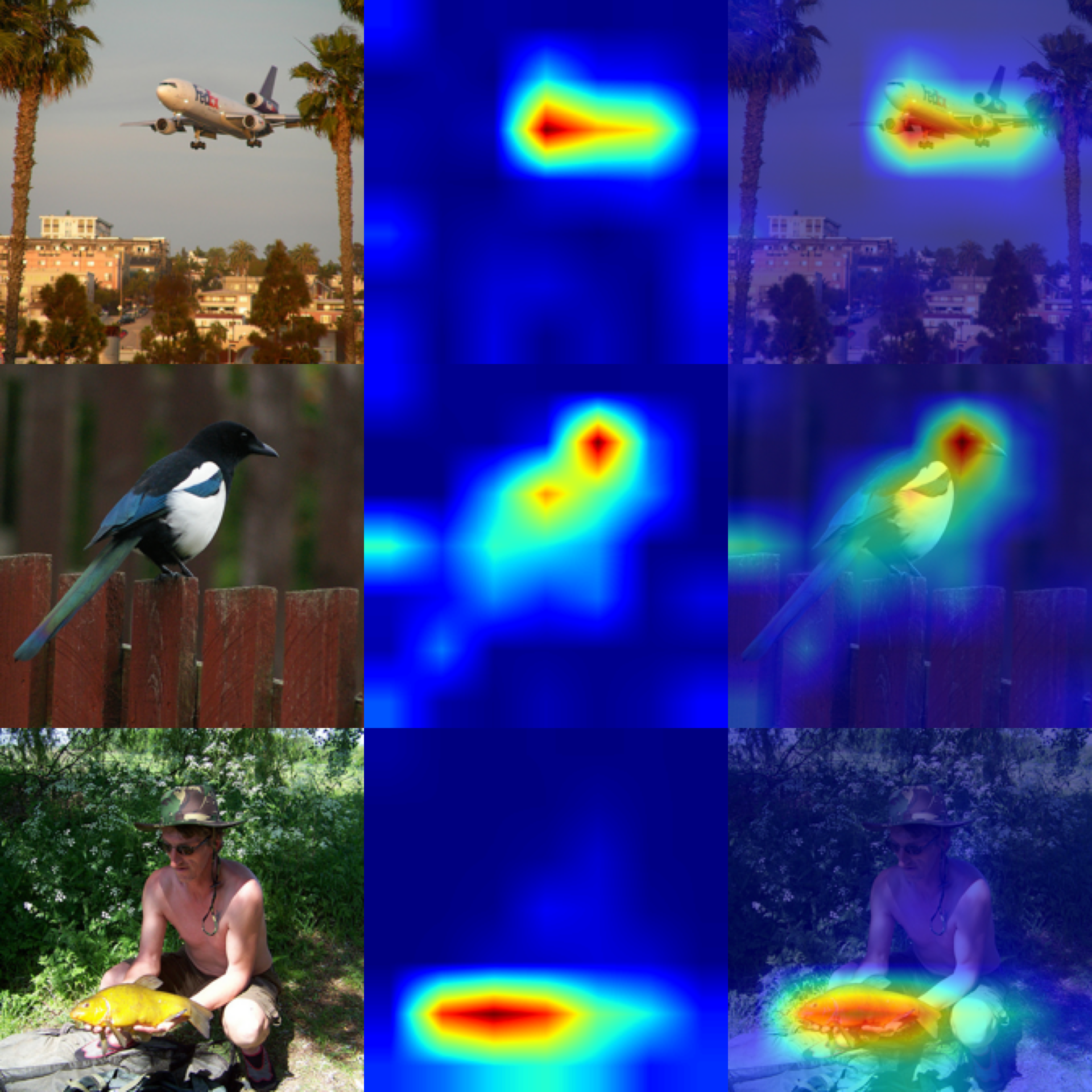}
    \caption{
    Visualizations of MambaVision's self-attention layers showing how the model learns to focus on semantically meaningful regions via attention maps (middle) and overlays (right).
    }
    \label{fig:attn_map}
\end{figure}

\begin{table}
\small
    \caption{Ablation study of on the effectiveness of different hybrid integration patterns. S and M denote self-attention and MambaVision token mixer blocks, respectively.}
    \label{tab:abl2}
\centering
\resizebox{.78\linewidth}{!}{
\setlength{\tabcolsep}{2.5pt}
  \begin{tabular}{lccc}
    \toprule
    Model  & Pattern & Params (M) &Top-1 \\
    \midrule	 
    Random  & - &31.8&81.3 \\
    First $N/2$ layers &SSSSMMMM  &31.8& 81.5 \\
    Mixed layers-1 &SMSMSMSM  &31.8& 81.4 \\
    Mixed layers-2 &MSMSMSMS  &31.8& 81.6 \\
    Last $N/4$ layers &MMMMMMSS  &31.8& 81.9 \\
    \hline
\rowcolor{Gray}
    \textbf{Last $N/2$ layers} &\textbf{MMMMSSSS}  &31.8& \textbf{82.3}\\
    \bottomrule
  \end{tabular} 
  }
\end{table}

\paragraph{Interpretability} To better understand how MambaVision processes visual information, we visualize the attention maps from the self-attention layers in the final stages. As shown in Fig.~\ref{fig:attn_map}, these visualizations reveal that the model learns to focus on semantically meaningful regions without explicit supervision. In the aircraft example, the attention clearly highlights the entire plane body, suggesting effective capture of object boundaries. For the bird image, we observe concentrated attention on distinctive features like the head and tail regions, demonstrating the model's ability to identify fine-grained details. In the case of object-human interaction (bottom row), the attention map shows strong activation on both the subject and the object being held, indicating that the self-attention layers successfully model relationships between different elements in the scene. These visualizations support our architectural design choice of using self-attention blocks in the final stages to capture global context and long-range dependencies.

\section{Conclusion}
In this work, we introduced MambaVision which is the first Mamba-Transformer hybrid backbone specifically tailored for vision applications. We proposed re-design of Mamba formulation to enhance global context representation learning capability. MambaVision achieves a new SOTA Pareto front in terms of Top-1 accuracy and image throughput, outperforming Transformer and Mamba-based models by a significant margin. Through extensive experimentation across multiple vision tasks, including classification, detection and segmentation, we demonstrated the versatility and effectiveness of our approach. Our systematic analysis of integration patterns revealed that positioning self-attention blocks in the final layers significantly improves the model's ability to capture long-range dependencies while maintaining efficiency. Furthermore, we successfully scaled MambaVision to ImageNet-21K pretraining, achieving strong performance that matches SOTA models, demonstrating its potential for large-scale vision applications. The success of MambaVision in addressing the limitations of pure Mamba-based architectures while leveraging their strengths opens new possibilities for vision backbone design. We hope these findings could be the foundation for a new class of hybrid vision models.

{
    \small
    \bibliographystyle{ieeenat_fullname}
    \bibliography{main}

\begin{thebibliography}{40}
\providecommand{\natexlab}[1]{#1}
\providecommand{\url}[1]{\texttt{#1}}
\expandafter\ifx\csname urlstyle\endcsname\relax
  \providecommand{\doi}[1]{doi: #1}\else
  \providecommand{\doi}{doi: \begingroup \urlstyle{rm}\Url}\fi

\bibitem[Chen et~al.(2021{\natexlab{a}})Chen, Fan, and Panda]{chen2021crossvit}
Chun-Fu~Richard Chen, Quanfu Fan, and Rameswar Panda.
\newblock Crossvit: Cross-attention multi-scale vision transformer for image classification.
\newblock In \emph{Proceedings of the IEEE/CVF international conference on computer vision}, pages 357--366, 2021{\natexlab{a}}.

\bibitem[Chen et~al.(2021{\natexlab{b}})Chen, Xie, Niu, Liu, Wei, and Tian]{chen2021visformer}
Zhengsu Chen, Lingxi Xie, Jianwei Niu, Xuefeng Liu, Longhui Wei, and Qi Tian.
\newblock Visformer: The vision-friendly transformer.
\newblock In \emph{Proceedings of the IEEE/CVF International Conference on Computer Vision}, pages 589--598, 2021{\natexlab{b}}.

\bibitem[Chu et~al.(2021)Chu, Tian, Wang, Zhang, Ren, Wei, Xia, and Shen]{chu2021twins}
Xiangxiang Chu, Zhi Tian, Yuqing Wang, Bo Zhang, Haibing Ren, Xiaolin Wei, Huaxia Xia, and Chunhua Shen.
\newblock Twins: Revisiting the design of spatial attention in vision transformers.
\newblock \emph{Advances in Neural Information Processing Systems}, 34, 2021.

\bibitem[Deng et~al.(2009)Deng, Dong, Socher, Li, Li, and Fei-Fei]{deng2009imagenet}
Jia Deng, Wei Dong, Richard Socher, Li-Jia Li, Kai Li, and Li Fei-Fei.
\newblock Imagenet: A large-scale hierarchical image database.
\newblock In \emph{2009 IEEE conference on computer vision and pattern recognition}, pages 248--255. Ieee, 2009.

\bibitem[Dosovitskiy et~al.(2020)Dosovitskiy, Beyer, Kolesnikov, Weissenborn, Zhai, Unterthiner, Dehghani, Minderer, Heigold, Gelly, et~al.]{dosovitskiy2020image}
Alexey Dosovitskiy, Lucas Beyer, Alexander Kolesnikov, Dirk Weissenborn, Xiaohua Zhai, Thomas Unterthiner, Mostafa Dehghani, Matthias Minderer, Georg Heigold, Sylvain Gelly, et~al.
\newblock An image is worth 16x16 words: Transformers for image recognition at scale.
\newblock In \emph{International Conference on Learning Representations}, 2020.

\bibitem[Elfwing et~al.(2018)Elfwing, Uchibe, and Doya]{elfwing2018sigmoid}
Stefan Elfwing, Eiji Uchibe, and Kenji Doya.
\newblock Sigmoid-weighted linear units for neural network function approximation in reinforcement learning.
\newblock \emph{Neural networks}, 107:\penalty0 3--11, 2018.

\bibitem[Gu and Dao(2023)]{gu2023mamba}
Albert Gu and Tri Dao.
\newblock Mamba: Linear-time sequence modeling with selective state spaces.
\newblock \emph{arXiv preprint arXiv:2312.00752}, 2023.

\bibitem[Gu et~al.(2021)Gu, Johnson, Goel, Saab, Dao, Rudra, and R{\'e}]{gu2021combining}
Albert Gu, Isys Johnson, Karan Goel, Khaled Saab, Tri Dao, Atri Rudra, and Christopher R{\'e}.
\newblock Combining recurrent, convolutional, and continuous-time models with linear state space layers.
\newblock \emph{Advances in neural information processing systems}, 34:\penalty0 572--585, 2021.

\bibitem[Han et~al.(2021)Han, Xiao, Wu, Guo, Xu, and Wang]{han2021transformer}
Kai Han, An Xiao, Enhua Wu, Jianyuan Guo, Chunjing Xu, and Yunhe Wang.
\newblock Transformer in transformer.
\newblock \emph{Advances in Neural Information Processing Systems}, 34:\penalty0 15908--15919, 2021.

\bibitem[Hatamizadeh et~al.(2023{\natexlab{a}})Hatamizadeh, Heinrich, Yin, Tao, Alvarez, Kautz, and Molchanov]{hatamizadeh2023fastervit}
Ali Hatamizadeh, Greg Heinrich, Hongxu Yin, Andrew Tao, Jose~M Alvarez, Jan Kautz, and Pavlo Molchanov.
\newblock Fastervit: Fast vision transformers with hierarchical attention.
\newblock \emph{arXiv preprint arXiv:2306.06189}, 2023{\natexlab{a}}.

\bibitem[Hatamizadeh et~al.(2023{\natexlab{b}})Hatamizadeh, Yin, Heinrich, Kautz, and Molchanov]{hatamizadeh2023global}
Ali Hatamizadeh, Hongxu Yin, Greg Heinrich, Jan Kautz, and Pavlo Molchanov.
\newblock Global context vision transformers.
\newblock In \emph{International Conference on Machine Learning}, pages 12633--12646. PMLR, 2023{\natexlab{b}}.

\bibitem[He et~al.(2016)He, Zhang, Ren, and Sun]{he2016deep}
Kaiming He, Xiangyu Zhang, Shaoqing Ren, and Jian Sun.
\newblock Deep residual learning for image recognition.
\newblock In \emph{Proceedings of the IEEE conference on computer vision and pattern recognition}, pages 770--778, 2016.

\bibitem[He et~al.(2017)He, Gkioxari, Doll{\'a}r, and Girshick]{he2017mask}
Kaiming He, Georgia Gkioxari, Piotr Doll{\'a}r, and Ross Girshick.
\newblock Mask r-cnn.
\newblock In \emph{Proceedings of the IEEE international conference on computer vision}, pages 2961--2969, 2017.

\bibitem[Hendrycks and Gimpel(2016)]{hendrycks2016gaussian}
Dan Hendrycks and Kevin Gimpel.
\newblock Gaussian error linear units (gelus).
\newblock \emph{arXiv preprint arXiv:1606.08415}, 2016.

\bibitem[Ioffe and Szegedy(2015)]{ioffe2015batch}
Sergey Ioffe and Christian Szegedy.
\newblock Batch normalization: Accelerating deep network training by reducing internal covariate shift.
\newblock In \emph{International conference on machine learning}, pages 448--456. PMLR, 2015.

\bibitem[Krizhevsky et~al.(2012)Krizhevsky, Sutskever, and Hinton]{krizhevsky2012imagenet}
Alex Krizhevsky, Ilya Sutskever, and Geoffrey~E Hinton.
\newblock Imagenet classification with deep convolutional neural networks.
\newblock In \emph{Advances in neural information processing systems}, pages 1097--1105, 2012.

\bibitem[Li et~al.(2022{\natexlab{a}})Li, Xia, Li, Li, Wang, Xiao, Wang, Zheng, and Pan]{li2022next}
Jiashi Li, Xin Xia, Wei Li, Huixia Li, Xing Wang, Xuefeng Xiao, Rui Wang, Min Zheng, and Xin Pan.
\newblock Next-vit: Next generation vision transformer for efficient deployment in realistic industrial scenarios.
\newblock \emph{arXiv preprint arXiv:2207.05501}, 2022{\natexlab{a}}.

\bibitem[Li et~al.(2022{\natexlab{b}})Li, Yuan, Wen, Hu, Evangelidis, Tulyakov, Wang, and Ren]{li2022efficientformer}
Yanyu Li, Geng Yuan, Yang Wen, Ju Hu, Georgios Evangelidis, Sergey Tulyakov, Yanzhi Wang, and Jian Ren.
\newblock Efficientformer: Vision transformers at mobilenet speed.
\newblock \emph{Advances in Neural Information Processing Systems}, 35:\penalty0 12934--12949, 2022{\natexlab{b}}.

\bibitem[Lin et~al.(2014)Lin, Maire, Belongie, Hays, Perona, Ramanan, Doll{\'a}r, and Zitnick]{lin2014microsoft}
Tsung-Yi Lin, Michael Maire, Serge Belongie, James Hays, Pietro Perona, Deva Ramanan, Piotr Doll{\'a}r, and C~Lawrence Zitnick.
\newblock Microsoft {COCO}: Common objects in context.
\newblock In \emph{ECCV}, 2014.

\bibitem[Liu et~al.(2024)Liu, Tian, Zhao, Yu, Xie, Wang, Ye, and Liu]{liu2024vmamba}
Yue Liu, Yunjie Tian, Yuzhong Zhao, Hongtian Yu, Lingxi Xie, Yaowei Wang, Qixiang Ye, and Yunfan Liu.
\newblock Vmamba: Visual state space model.
\newblock \emph{arXiv preprint arXiv:2401.10166}, 2024.

\bibitem[Liu et~al.(2021)Liu, Lin, Cao, Hu, Wei, Zhang, Lin, and Guo]{liu2021swin}
Ze Liu, Yutong Lin, Yue Cao, Han Hu, Yixuan Wei, Zheng Zhang, Stephen Lin, and Baining Guo.
\newblock Swin transformer: Hierarchical vision transformer using shifted windows.
\newblock In \emph{Proceedings of the IEEE/CVF International Conference on Computer Vision}, pages 10012--10022, 2021.

\bibitem[Liu et~al.(2022{\natexlab{a}})Liu, Hu, Lin, Yao, Xie, Wei, Ning, Cao, Zhang, Dong, et~al.]{liu2022swin}
Ze Liu, Han Hu, Yutong Lin, Zhuliang Yao, Zhenda Xie, Yixuan Wei, Jia Ning, Yue Cao, Zheng Zhang, Li Dong, et~al.
\newblock Swin transformer v2: Scaling up capacity and resolution.
\newblock In \emph{Proceedings of the IEEE/CVF Conference on Computer Vision and Pattern Recognition}, pages 12009--12019, 2022{\natexlab{a}}.

\bibitem[Liu et~al.(2022{\natexlab{b}})Liu, Mao, Wu, Feichtenhofer, Darrell, and Xie]{liu2022convnet}
Zhuang Liu, Hanzi Mao, Chao-Yuan Wu, Christoph Feichtenhofer, Trevor Darrell, and Saining Xie.
\newblock A convnet for the 2020s.
\newblock In \emph{Proceedings of the IEEE/CVF Conference on Computer Vision and Pattern Recognition}, pages 11976--11986, 2022{\natexlab{b}}.

\bibitem[Patro and Agneeswaran(2024)]{patro2024simba}
Badri~N Patro and Vijay~S Agneeswaran.
\newblock Simba: Simplified mamba-based architecture for vision and multivariate time series.
\newblock \emph{arXiv preprint arXiv:2403.15360}, 2024.

\bibitem[Pei et~al.(2024)Pei, Huang, and Xu]{pei2024efficientvmamba}
Xiaohuan Pei, Tao Huang, and Chang Xu.
\newblock Efficientvmamba: Atrous selective scan for light weight visual mamba.
\newblock \emph{arXiv preprint arXiv:2403.09977}, 2024.

\bibitem[Radosavovic et~al.(2020)Radosavovic, Kosaraju, Girshick, He, and Doll{\'a}r]{radosavovic2020designing}
Ilija Radosavovic, Raj~Prateek Kosaraju, Ross Girshick, Kaiming He, and Piotr Doll{\'a}r.
\newblock Designing network design spaces.
\newblock In \emph{Proceedings of the IEEE/CVF conference on computer vision and pattern recognition}, pages 10428--10436, 2020.

\bibitem[Tan and Le(2021)]{tan2021efficientnetv2}
Mingxing Tan and Quoc Le.
\newblock Efficientnetv2: Smaller models and faster training.
\newblock In \emph{International Conference on Machine Learning}, pages 10096--10106. PMLR, 2021.

\bibitem[Touvron et~al.(2021)Touvron, Cord, Douze, Massa, Sablayrolles, and J{\'e}gou]{touvron2021training}
Hugo Touvron, Matthieu Cord, Matthijs Douze, Francisco Massa, Alexandre Sablayrolles, and Herv{\'e} J{\'e}gou.
\newblock Training data-efficient image transformers \& distillation through attention.
\newblock In \emph{International Conference on Machine Learning}, pages 10347--10357. PMLR, 2021.

\bibitem[Touvron et~al.(2022)Touvron, Cord, and J{\'e}gou]{touvron2022deit}
Hugo Touvron, Matthieu Cord, and Herv{\'e} J{\'e}gou.
\newblock Deit iii: Revenge of the vit.
\newblock In \emph{Computer Vision--ECCV 2022: 17th European Conference, Tel Aviv, Israel, October 23--27, 2022, Proceedings, Part XXIV}, pages 516--533. Springer, 2022.

\bibitem[Tu et~al.(2022)Tu, Talebi, Zhang, Yang, Milanfar, Bovik, and Li]{tu2022maxvit}
Zhengzhong Tu, Hossein Talebi, Han Zhang, Feng Yang, Peyman Milanfar, Alan Bovik, and Yinxiao Li.
\newblock Maxvit: Multi-axis vision transformer.
\newblock In \emph{Computer Vision--ECCV 2022: 17th European Conference, Tel Aviv, Israel, October 23--27, 2022, Proceedings, Part XXIV}, pages 459--479. Springer, 2022.

\bibitem[Vaswani et~al.(2017)Vaswani, Shazeer, Parmar, Uszkoreit, Jones, Gomez, Kaiser, and Polosukhin]{vaswani2017attention}
Ashish Vaswani, Noam Shazeer, Niki Parmar, Jakob Uszkoreit, Llion Jones, Aidan~N Gomez, {\L}ukasz Kaiser, and Illia Polosukhin.
\newblock Attention is all you need.
\newblock In \emph{Advances in neural information processing systems}, pages 5998--6008, 2017.

\bibitem[Wang et~al.(2021)Wang, Xie, Li, Fan, Song, Liang, Lu, Luo, and Shao]{wang2021pyramid}
Wenhai Wang, Enze Xie, Xiang Li, Deng-Ping Fan, Kaitao Song, Ding Liang, Tong Lu, Ping Luo, and Ling Shao.
\newblock Pyramid vision transformer: A versatile backbone for dense prediction without convolutions.
\newblock In \emph{Proceedings of the IEEE/CVF International Conference on Computer Vision}, pages 568--578, 2021.

\bibitem[Wightman et~al.(2021)Wightman, Touvron, and J{\'e}gou]{wightman2021resnet}
Ross Wightman, Hugo Touvron, and Herv{\'e} J{\'e}gou.
\newblock Resnet strikes back: An improved training procedure in timm.
\newblock \emph{arXiv preprint arXiv:2110.00476}, 2021.

\bibitem[Xiao et~al.(2018)Xiao, Liu, Zhou, Jiang, and Sun]{xiao2018unified}
Tete Xiao, Yingcheng Liu, Bolei Zhou, Yuning Jiang, and Jian Sun.
\newblock Unified perceptual parsing for scene understanding.
\newblock In \emph{Proceedings of the European Conference on Computer Vision (ECCV)}, pages 418--434, 2018.

\bibitem[Xie et~al.(2017)Xie, Girshick, Doll{\'a}r, Tu, and He]{xie2017aggregated}
Saining Xie, Ross Girshick, Piotr Doll{\'a}r, Zhuowen Tu, and Kaiming He.
\newblock Aggregated residual transformations for deep neural networks.
\newblock In \emph{Proceedings of the IEEE conference on computer vision and pattern recognition}, pages 1492--1500, 2017.

\bibitem[Xu et~al.(2021)Xu, Xu, Chang, and Tu]{xu2021co}
Weijian Xu, Yifan Xu, Tyler Chang, and Zhuowen Tu.
\newblock Co-scale conv-attentional image transformers.
\newblock In \emph{Proceedings of the IEEE/CVF International Conference on Computer Vision}, pages 9981--9990, 2021.

\bibitem[Yang et~al.(2021)Yang, Li, Zhang, Dai, Xiao, Yuan, and Gao]{yang2021focal}
Jianwei Yang, Chunyuan Li, Pengchuan Zhang, Xiyang Dai, Bin Xiao, Lu Yuan, and Jianfeng Gao.
\newblock Focal attention for long-range interactions in vision transformers.
\newblock \emph{Advances in Neural Information Processing Systems}, 34, 2021.

\bibitem[Yu et~al.(2022)Yu, Luo, Zhou, Si, Zhou, Wang, Feng, and Yan]{yu2022metaformer}
Weihao Yu, Mi Luo, Pan Zhou, Chenyang Si, Yichen Zhou, Xinchao Wang, Jiashi Feng, and Shuicheng Yan.
\newblock Metaformer is actually what you need for vision.
\newblock In \emph{Proceedings of the IEEE/CVF Conference on Computer Vision and Pattern Recognition}, pages 10819--10829, 2022.

\bibitem[Zhou et~al.(2017)Zhou, Zhao, Puig, Fidler, Barriuso, and Torralba]{zhou2017scene}
Bolei Zhou, Hang Zhao, Xavier Puig, Sanja Fidler, Adela Barriuso, and Antonio Torralba.
\newblock Scene parsing through ade20k dataset.
\newblock In \emph{Proceedings of the IEEE conference on computer vision and pattern recognition}, pages 633--641, 2017.

\bibitem[Zhu et~al.(2024)Zhu, Liao, Zhang, Wang, Liu, and Wang]{zhu2024vision}
Lianghui Zhu, Bencheng Liao, Qian Zhang, Xinlong Wang, Wenyu Liu, and Xinggang Wang.
\newblock Vision mamba: Efficient visual representation learning with bidirectional state space model.
\newblock In \emph{Proceedings of the 41st International Conference on Machine Learning}, pages 62429--62442. PMLR, 2024.

\end{thebibliography}
}

\newpage
\section*{Appendix}

\renewcommand{\thesection}{\Alph{section}}
\renewcommand\thefigure{S.\arabic{figure}}
\setcounter{figure}{0}
\renewcommand\thetable{S.\arabic{table}}
\setcounter{table}{0}

\begin{figure*}[t]
\centering
\includegraphics[width=\linewidth]{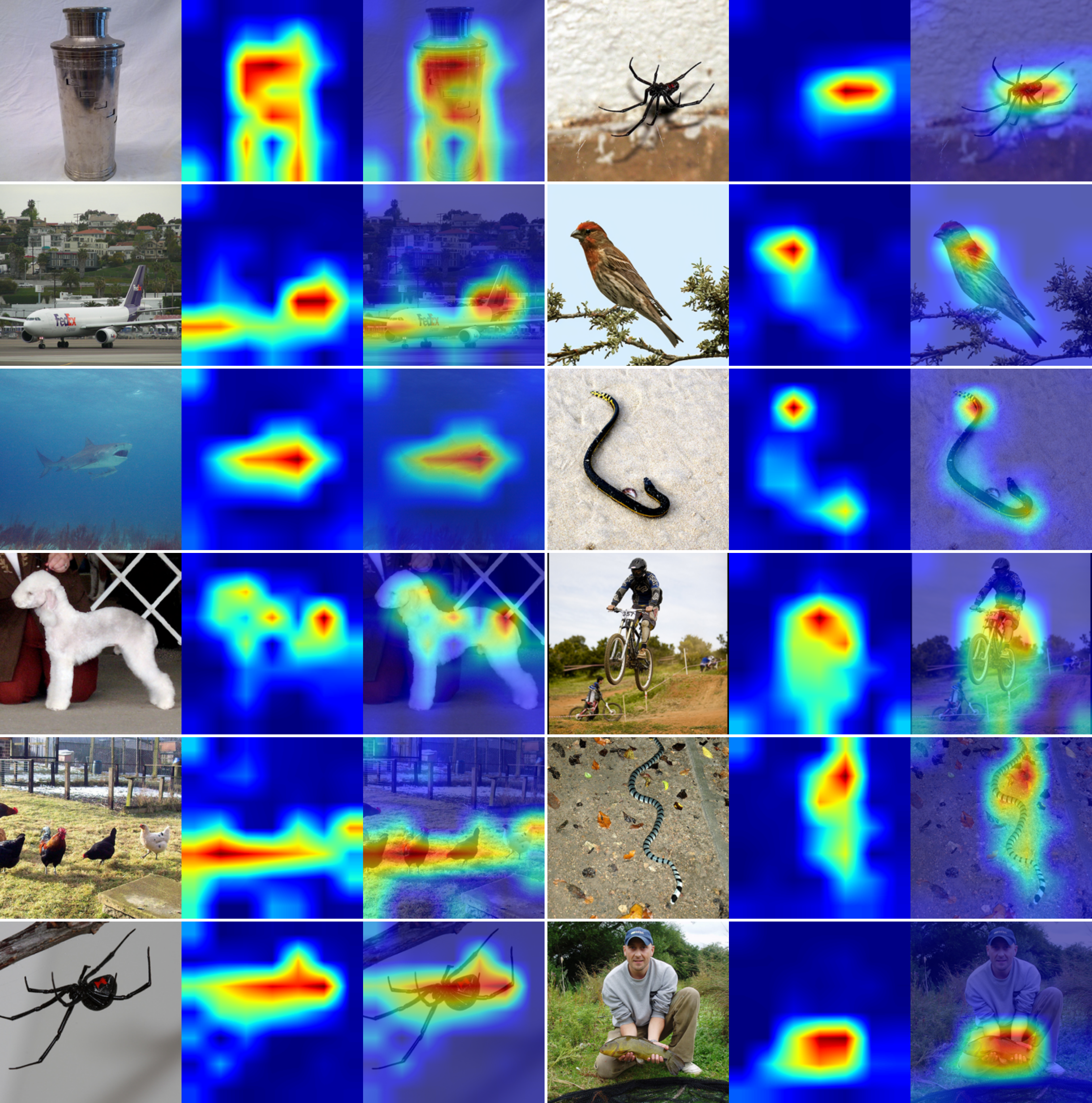}
\caption{Visualization of MambaVision's attention patterns. Each row contains two example cases, with each case showing a triplet of: (left) original input image, (middle) attention heat map, and (right) attention overlay on the input image. The examples showcase diverse scenarios: containers and spiders (row 1), aircraft and birds (row 2), marine life and snakes (row 3), groomed dogs and extreme sports (row 4), poultry and snakes (row 5), and arachnids and outdoor activities (row 6). The attention maps reveal how MambaVision effectively localizes key semantic regions and object boundaries across this wide range of categories.}
\label{fig:attention_viz}
\end{figure*}

\section{Ablation Study}
To determine the optimal window size for MambaVision models, we study its impact on the performance of MambaVision-T in different tasks such as image classification, object detection and instance segmentation. Given $Q, K, V$ as the query, key and value tensors respectively, self-attention is computed according to
\begin{equation}
    {\rm Attention}({Q}, {K}, {V}) = {\rm Softmax}(\frac{{Q}{K}^\mathsf{T}}{\sqrt{d_{h}}}){V}.
    \label{eqn:mhsa_abl}
\end{equation}
$d_{h}$ represents the number of attention heads. If the input size is larger than the window size, the attention is computed in the local windows. Specifically, we study two different architectures with window sizes of 7 and 14 in their stage 3 of the model. We also measure image throughput for the task of image classification with a batch size of 128. As presented in Table~\ref{tab:abl_study_window}, our analysis reveals that increasing the window size to 14 offers a favorable trade-off between performance and computational cost. While maintaining nearly identical throughput (6298 img/s vs. 6318 img/s), the larger window size achieves consistent improvements across all vision benchmarks: ImageNet top-1 accuracy increases to 82.3\%, COCO mask AP improves to 41.8\%. These gains, though modest, come with minimal computational overhead on modern hardware such as the NVIDIA A100 GPU. Based on this empirical evidence, we selected 14 and 7 as our default window sizes, as this combination provides better vision understanding capabilities while preserving the model's efficiency. The negligible 0.3\% decrease in throughput is well justified by the improved performance in various vision tasks. 

\begin{table}[h]
\centering
\resizebox{\columnwidth}{!}{%
\setlength{\tabcolsep}{4pt}
\begin{tabular}{l|c|c|c|cc}
\Xhline{1.0pt}
\multirow{2}{*}{Model} & \multirow{2}{*}{\begin{tabular}[c]{@{}c@{}}Window\\ Size\end{tabular}} & \multirow{2}{*}{\begin{tabular}[c]{@{}c@{}}Throughput\\ (img/s)\end{tabular}} & \multicolumn{1}{c|}{ImageNet} & \multicolumn{2}{c}{COCO} \\
 & & & top-1 & AP$^\text{box}$ & AP$^\text{mask}$ \\
\hline
MambaVision-T & 7,7 & 6318 & 82.2 & 46.4 & 41.7 \\
MambaVision-T & 14,7 & 6298 & 82.3 & 46.4 & 41.8 \\
\Xhline{1.0pt}
\end{tabular}%
}
\caption{Ablation study on window size for MambaVision model's performance. Experiments on COCO dataset~\cite{lin2014microsoft} are performed using Mask-RCNN~\cite{he2017mask} head and ×1 LR schedule. Throughput is measured for image classification on a single NVIDIA A100 GPU with batch size 128.}
\label{tab:abl_study_window}
\end{table}

\section{Architecture Details}
\label{sec:arch}
In Table~\ref{table:arch-spec-abl}, we present the comprehensive architectural specifications of MambaVision variants. The backbone follows a hierarchical design with 4 stages, each employing convolutional down-sampling operations that progressively reduce spatial resolution by a factor of two. A key innovation in our architecture appears in Stages 3 and 4, where we introduce a hybrid design that synergistically combines Mamba-based sequence modeling with self-attention mechanisms. This hybrid approach leverages Mamba's efficient sequence processing capabilities while benefiting from the global context modeling strengths of self-attention layers. Each variant (T, S, B, and L) maintains this fundamental structure while scaling the channel dimensions and layer counts to achieve different complexity-performance trade-offs.

\begin{table*}[t]
\small
\centering
\addtolength{\tabcolsep}{-2pt}
\resizebox{1.0\linewidth}{!}{
\begin{tabular}{c|c|c|c|c|c}
 & \begin{tabular}[c]{@{}c@{}}Output Size \\ (Downs. Rate)\end{tabular} & MambaVision-T  &MambaVision-S & MambaVision-B &  MambaVision-L \\
\hline
\hline
\multirow{3}{*}{Stem} & \multirow{3}{*}{\begin{tabular}[c]{@{}c@{}}112$\times$112\\ (2$\times$)\end{tabular}} & $\begin{bmatrix}\text{Conv-BN-ReLU}\\\text{C:32, S:2}\end{bmatrix}$ $\times$ 1  & $\begin{bmatrix}\text{Conv-BN-ReLU}\\\text{C:64, S:2}\end{bmatrix}$ $\times$ 1  &$\begin{bmatrix}\text{Conv-BN-ReLU}\\\text{C:64, S:2}\end{bmatrix}$ $\times$ 1   & $\begin{bmatrix}\text{Conv-BN-ReLU}\\\text{C:64, S:2}\end{bmatrix}$ $\times$ 1  \\
\cline{3-6}
& & $\begin{bmatrix}\text{Conv-BN-ReLU}\\\text{C:80}\end{bmatrix}$ $\times$ 1   & $\begin{bmatrix}\text{Conv-BN-ReLU}\\\text{C:96}\end{bmatrix}$ $\times$ 1    & $\begin{bmatrix}\text{Conv-BN-ReLU}\\\text{C:128}\end{bmatrix}$ $\times$ 1   & $\begin{bmatrix}\text{Conv-BN-ReLU}\\\text{C:196}\end{bmatrix}$ $\times$ 1   \\
\hline
\multirow{3}{*}{Stage 1} & \multirow{3}{*}{\begin{tabular}[c]{@{}c@{}}56$\times$56\\ (4$\times$)\end{tabular}} & Conv, C:160, S:2  & Conv, C:192, S:2  & Conv, C:256, S:2 & Conv, C:392, S:2  \\
\cline{3-6}
& & $\begin{bmatrix}\text{ResBlock}\\\text{C:160}\end{bmatrix}$ $\times$1,  & $\begin{bmatrix}\text{ResBlock}\\\text{C:192}\end{bmatrix}$ $\times$ 3,   & $\begin{bmatrix}\text{ResBlock}\\\text{C:256}\end{bmatrix}$ $\times$ 3,   & $\begin{bmatrix}\text{ResBlock}\\\text{C:392}\end{bmatrix}$ $\times$ 3,   \\
\hline
\multirow{3}{*}{Stage 2} & \multirow{3}{*}{\begin{tabular}[c]{@{}c@{}}28$\times$28\\ (8$\times$)\end{tabular}} & Conv, C:320, S:2  & Conv, C:384, S:2  & Conv, C:512, S:2  & Conv, C:768, S:2  \\
\cline{3-6}
& & $\begin{bmatrix}\text{ResBlock}\\\text{C:320}\end{bmatrix}$ $\times$ 3,   & $\begin{bmatrix}\text{ResBlock}\\\text{C:384}\end{bmatrix}$ $\times$ 3,    & $\begin{bmatrix}\text{ResBlock}\\\text{C:512}\end{bmatrix}$ $\times$ 3,   & $\begin{bmatrix}\text{ResBlock}\\\text{C:768}\end{bmatrix}$ $\times$ 3,   \\
\hline
\multirow{3}{*}{Stage 3} & \multirow{3}{*}{\begin{tabular}[c]{@{}c@{}}14$\times$14\\ (16$\times$)\end{tabular}} & Conv, C:640, S:2 & Conv, C:768, S:2  &Conv, C:1024, S:2   &Conv, C:1568, S:2 \\
\cline{3-6}
& & $\begin{bmatrix}\text{MV}\\\text{C:640}\end{bmatrix}$ $\times$ 4, $\begin{bmatrix}\text{SA}\\\text{C:640, head:8}\end{bmatrix}$ $\times$ 4 ,  & $\begin{bmatrix}\text{MV}\\\text{C:768}\end{bmatrix}$ $\times$ 4,$\begin{bmatrix}\text{SA}\\\text{C:768, head:8}\end{bmatrix}$ $\times$ 3,    & $\begin{bmatrix}\text{MV}\\\text{C:1024}\end{bmatrix}$ $\times$ 4,$\begin{bmatrix}\text{SA}\\\text{C:1024, head:8}\end{bmatrix}$ $\times$ 4,   & $\begin{bmatrix}\text{MV}\\\text{C:1568}\end{bmatrix}$ $\times$ 4,$\begin{bmatrix}\text{SA}\\\text{C:1568, head:16}\end{bmatrix}$ $\times$ 3,   \\
\hline
\multirow{3}{*}{Stage 4} & \multirow{3}{*}{\begin{tabular}[c]{@{}c@{}}7$\times$7\\ (32$\times$)\end{tabular}} & Conv, C:1280, S:2  & Conv, C:1536, S:2 & Conv, C:2048, S:2  &  Conv, C:3136, S:2  \\
\cline{3-6}
& & $\begin{bmatrix}\text{MV}\\\text{C:1280}\end{bmatrix}$ $\times$ 4,$\begin{bmatrix}\text{SA}\\\text{C:1280, head:16}\end{bmatrix}$ $\times$ 4,   & $\begin{bmatrix}\text{MV}\\\text{C:1536}\end{bmatrix}$ $\times$ 4,$\begin{bmatrix}\text{SA}\\\text{C:1536, head:16}\end{bmatrix}$ $\times$ 2,    & $\begin{bmatrix}\text{MV}\\\text{C:2048}\end{bmatrix}$ $\times$ 4,$\begin{bmatrix}\text{SA}\\\text{C:2048, head:16}\end{bmatrix}$ $\times$ 2,   & $\begin{bmatrix}\text{MV}\\\text{C:3136}\end{bmatrix}$ $\times$ 4,$\begin{bmatrix}\text{SA}\\\text{C:3136, head:32}\end{bmatrix}$ $\times$ 2,   \\
\end{tabular}
}
\normalsize
\caption{Architecture configurations of MambaVision models. SA and MV refer to self-attention and MambaVision mixer blocks respectively. BN denote Batch Normalization.}
\label{table:arch-spec-abl}
\end{table*}

\section{Training Details}
Image classification experiments are conducted on the ImageNet-1K dataset~\cite{deng2009imagenet}. All models have been trained for 300 epochs using 32 A100 GPUs, with LAMB optimizer, batch size of 4096, and learning rate of 4e-3. The self-attention formulation in stages 3 and 4 of all MambaVision variants use a window size of 14 and 7, respectively. To evaluate the performance of downstream tasks, we used our pre-trained models as backbones for object detection, instance segmentation, and semantic segmentation tasks using the MS COCO dataset~\cite{lin2014microsoft} and ADE20K dataset~\cite{zhou2017scene}, respectively. For all downstream tasks, we used an AdamW optimizer and batch size of 16. Specifically, for object detection and instance segmentation, we used the Cascade Mask-RCNN~\cite{he2017mask} head with hyperparameters such as ×3 LR schedule. For semantic segmentation, we used a UperNet network~\cite{xiao2018unified} segmentation head.

\section{Interpretability}
\label{sec:interop}

To demonstrate the interpretability of MambaVision models, we visualize the attention patterns learned by our model across diverse object categories. Figure~\ref{fig:attention_viz} presents a comprehensive analysis of attention mechanisms through paired examples.

Our paired visualization analysis reveals several key insights about MambaVision's visual processing capabilities:

\begin{itemize}
    \item \textbf{Consistent Pattern Recognition}: Each triplet (input-heatmap-overlay) demonstrates how the model maintains consistent attention patterns across different instances of similar object categories.
    
    \item \textbf{Contextual Understanding}: The paired examples within each row often represent contrasting scenarios (e.g., man-made objects vs. natural subjects), showing the model's adaptability across domains.
    
    \item \textbf{Fine-grained Detail}: The attention heat maps precisely highlight discriminative features, from the texture of animal fur to the structural elements of vehicles and containers.
    
    \item \textbf{Robust Localization}: Across all example pairs, the overlaid visualizations demonstrate accurate object boundary detection, regardless of the subject's position or background complexity.
\end{itemize}

\end{document}